\documentclass[aps,pra,reprint]{revtex4-1}
\usepackage{amsmath,amssymb,amsthm,bm,graphicx,hyperref,listings,lmodern,physics,tikz,upgreek}
\usepackage[per-mode=fraction]{siunitx}
\usepackage[caption=false]{subfig}
\usetikzlibrary{decorations.pathmorphing}
\newcommand{\ce}{\ensuremath{\textrm{e}}}
\DeclareMathOperator*{\argmin}{arg\,min}
\newcommand{\KLD}[2]{D_{\mathrm{KL}} \left( \left. \left. #1 \right|\right| #2 \right) }
\newcommand\given[1][]{\:#1\vert\:}

\hypersetup{
	colorlinks=true,
	linkcolor=red,
	citecolor=blue,
}

\begin{document}
\title{Near-Term Quantum-Classical Associative Adversarial Networks}
\author{Eric R. \surname{Anschuetz}}
\email{eans@mit.edu}
\affiliation{Department of Physics, Massachusetts Institute of Technology, 77 Massachusetts Avenue, Cambridge, MA 02139, USA}
\author{Cristian \surname{Zanoci}}
\email{czanoci@mit.edu}
\affiliation{Department of Physics, Massachusetts Institute of Technology, 77 Massachusetts Avenue, Cambridge, MA 02139, USA}
\preprint{MIT-CTP/5125}

\begin{abstract}
We introduce a new hybrid quantum-classical adversarial machine learning architecture called a quantum-classical associative adversarial network (QAAN). This architecture consists of a classical generative adversarial network with a small auxiliary quantum Boltzmann machine that is simultaneously trained on an intermediate layer of the discriminator of the generative network. We numerically study the performance of QAANs compared to their classical counterparts on the MNIST and CIFAR-10 data sets, and show that QAANs attain a higher quality of learning when evaluated using the Inception score and the Fr\'{e}chet Inception distance. As the QAAN architecture only relies on sampling simple local observables of a small quantum Boltzmann machine, this model is particularly amenable for implementation on the current and next generations of quantum devices.
\end{abstract}

\maketitle

\section{Introduction}\label{sec:introduction}

Quantum mechanics is widely believed to produce distributions of data difficult to replicate classically~\cite{boixo2018characterizing}. As deep learning is implemented through applying a series of transformations to latent probability distributions to approximate empirical distributions given by data, this has led to a trend in recent years of applying quantum computing techniques to machine learning~\cite{biamonte2017quantum,wiebe2014quantum,farhi2018classification,Pudenz2013,neven2009training,lloyd2013quantum,denchev2012robust,killoran2018continuous,PhysRevA.98.042308}. One approach to this task consists of applying well-known quantum algorithms---such as the HHL algorithm~\cite{PhysRevLett.103.150502}---to replace their classical counterparts---such as the BLAS library~\cite{biamonte2017quantum}. This method has been applied to principal component analysis~\cite{lloyd2014quantum}, support vector machines~\cite{PhysRevLett.113.130503}, and most recently, generative adversarial networks~\cite{dallaire2018quantum,lloyd2018quantum}. These architectures, however, require large quantum networks and the extensive use of quantum memory, which are not yet available on the current generation of quantum devices~\cite{aaronson2015read}.

Instead, one could focus on translating such classical machine learning algorithms for use on current and near-term quantum devices, in the so-called Noisy Intermediate-Scale Quantum (NISQ) era~\cite{Preskill2018quantumcomputingin}. This has been made possible by recent developments in neutral atom and ion trap architectures~\cite{Endresaah3752, zhang2017observation}, as well as gate-model processors~\cite{neill2018blueprint, kandala2017hardware}. Inspired by variational hybrid quantum-classical algorithms~\cite{farhi2014quantum,mcclean2016theory,peruzzo2014variational}, there has been an effort to create machine learning models that use small quantum devices to aid in the training of classical machine learning architectures. Recently studied hybrid architectures include generalizations of Helmholtz machines~\cite{benedetti2018quantum,perdomo2018opportunities} and variational autoencoders~\cite{khoshaman2018quantum}. In these models, the small quantum device encodes a quantum Boltzmann machine (QBM)~\cite{amin2016quantum}, which is a quantum generalization of classical restricted Boltzmann machines (RBMs)~\cite{smolensky1986information}. In this paper, we focus on using small QBMs to improve the performance of generative adversarial networks (GANs).

GANs are the state-of-the-art classical machine learning architectures for unsupervised learning~\cite{arjovsky2017wasserstein,goodfellow2016deep,goodfellow2016nips,gulrajani2017improved,kodali2018convergence,metz2016unrolled,NIPS2014_5423}. They have a wide range of applications, including regression and classification~\cite{radford2015unsupervised,NIPS2016_6125}, image generation~\cite{denton2015deep,reed2016generative,karras2018style}, and image-to-image translation~\cite{isola2017image,li2016precomputed}. GANs can be thought of as a zero-sum game between two players---the generator and the discriminator---each implemented as a neural network. Taking the concrete setting of image generation, the generator learns to create images resembling a given data set of authentic images, and the discriminator learns to distinguish between images produced by the generator and images drawn from the true data set~\cite{NIPS2014_5423}. The generator does not have direct access to the data set, and only learns how to create images through feedback from the discriminator---that is, through an error signal backpropagated through the GAN.

Although GANs are the most ubiquitous adversarial models, they are notoriously difficult to train~\cite{NIPS2016_6125}. One of the major reasons GANs are difficult to train is that the generator and discriminator learn at different rates generically. One approach towards combating this issue is through the use of associative adversarial networks (AANs)~\cite{arici2016associative}. In this architecture, a Boltzmann machine acts as an associative memory that learns the high-level feature distribution of a layer of the discriminator. The generator then draws samples from the distribution approximated by the Boltzmann machine. The associative memory also adds expresivity to the network by providing the generator with inputs drawn from a more meaningful, data-specific distribution, rather than a uniform or Gaussian distribution as is the case for standard GANs. Motivated by the observed improvement in performance of QBMs over RBMs~\cite{amin2016quantum,PhysRevA.96.062327}, we propose a method of implementing hybrid quantum-classical AANs (QAANs), where the associative memory is instead provided by a QBM. 
%We expect this change to leverage the quantum advantage of the quantum associative memory, resulting in better generated samples.

The structure of the paper is as follows. In Sec.~\ref{sec:class-advers-netw}, we give an introduction to classical adversarial networks and, in particular, AANs. In Sec.~\ref{sec:near-term-quantum}, we construct a quantum analog of AANs, and give a numerical comparison between classical and quantum AANs in Sec.~\ref{sec:numerical-results}. Finally, we discuss our results and future research directions in Sec.~\ref{sec:discussion}.

\section{Classical Generative Models}\label{sec:class-advers-netw}

\subsection{Restricted Boltzmann Machines}\label{sec:boltzmann-machines}
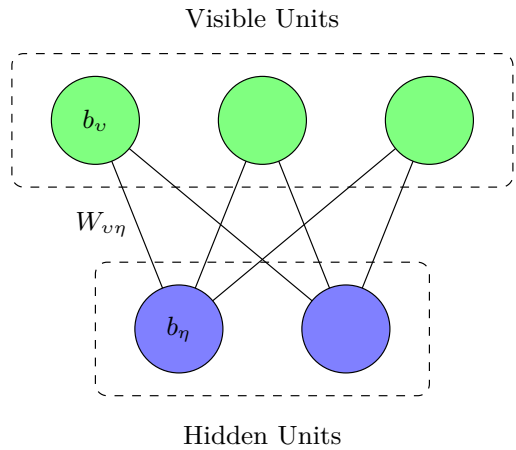
\begin{figure}
\begin{center}
\resizebox{0.8\linewidth}{!}{
\begin{tikzpicture}
\draw (-1, -5) -- node[anchor=east]{$W_{\upsilon\eta}$} (-2, -2.5);
\draw (-1, -5) -- (0, -2.5);
\draw (-1, -5) -- (2, -2.5);
\draw (1, -5) -- (-2, -2.5);
\draw (1, -5) -- (0, -2.5);
\draw (1, -5) -- (2, -2.5);
\node[draw,circle,fill=blue!50,minimum size=30,inner sep=0] at (-1, -5) {$b_{\eta}$};
\node[draw,circle,fill=blue!50,minimum size=30,inner sep=0] at (1, -5) {};
\draw[dashed,rounded corners] (-2, -5.8) rectangle (2, -4.2) {};
\node[anchor=south] at (0, -6.5) {Hidden Units};
\node[draw,circle,fill=green!50,minimum size=30,inner sep=0] at (-2, -2.5) {$b_{\upsilon}$};
\node[draw,circle,fill=green!50,minimum size=30,inner sep=0] at (0, -2.5) {};
\node[draw,circle,fill=green!50,minimum size=30,inner sep=0] at (2, -2.5) {};
\draw[dashed,rounded corners] (-3, -3.3) rectangle (3, -1.7) {};
\node[anchor=south] at (0, -1.5) {Visible Units};
\end{tikzpicture}
}
\caption{\label{fig:rbm}Schematic depiction of an RBM. The classical spin system with an energy given by Eq.~\eqref{rbmenergyrel} thermalizes and samples are drawn from the system's visible units (green). $W_{\upsilon\eta}$ terms represent interactions between visible and hidden units (blue), $b_\upsilon$ terms represent on-site interactions for the visible units, and $b_\eta$ terms represent on-site interactions for the hidden units. A QBM has an identical structure, but its Hamiltonian (Eq.~\eqref{qbmhamiltonianrel}) also includes terms not diagonal in the computational basis.}
\end{center}
\end{figure}

A \textit{Boltzmann machine} is an energy-based generative model, and one of the first neural networks capable of learning internal representations for and sampling from arbitrary probability distributions~\cite{hinton1986learning}. Recent work has seen successful applications of this model to a wide variety of machine learning tasks, including image~\cite{hinton2006fast}, text~\cite{salakhutdinov2009semantic}, and speech~\cite{mohamed2012acoustic} generation. It also serves as a key component in other machine learning architectures, such as deep belief networks~\cite{hinton2006fast}. A Boltzmann machine is characterized by an energy function
\begin{equation}\label{bmerel}
E\left(\bm{z};\bm{\theta}\right)=-\sum\limits_a b_a z_a-\sum\limits_{a<b}W_{ab}z_a z_b,
\end{equation}
where $\bm{z}\in\left\{0,1\right\}^N$ is a binary vector and $\bm{\theta} = \left\{\bm{b},\bm{W}\right\}$ are the model parameters. The model can be viewed as a two-layer network by dividing its nodes into \textit{visible units}, which represent the input data, and \textit{hidden units}, which form an internal representation of the data. In practice, since training a general Boltzmann machine is impractical, we consider \textit{restricted Boltzmann machines} (RBMs) which further simplify the model to only contain connections between visible and hidden units~\cite{smolensky1986information} (see Fig.~\ref{fig:rbm}). By labeling the indices of visible nodes by $\upsilon$ and the indices of hidden nodes by $\eta$, we can separate our vector as $\bm{z}=\left(\bm{v};\bm{h}\right)$, and rewrite the energy function in the form:
\begin{equation}
E\left(\bm{v}, \bm{h}; \bm{\theta}\right)=-\sum\limits_\upsilon b_\upsilon v_\upsilon-\sum\limits_\eta b_\eta h_\eta-\sum\limits_{\upsilon,\eta}W_{\upsilon\eta}v_\upsilon h_\eta.\label{rbmenergyrel}
\end{equation}
For an input vector $\bm{v}$, this network assigns the probability 
\begin{equation}
P\left({\bm{v}};\bm{\theta}\right)=\frac{1}{Z}\sum\limits_{\bm{h}}\ce^{-E\left(\bm{v}, \bm{h}; \bm{\theta}\right)},
\end{equation}
where $Z = \sum\limits_{\bm{v}}\sum\limits_{\bm{h}}\ce^{-E\left(\bm{v}, \bm{h}; \bm{\theta}\right)}$ is the partition function. The model parameters $\bm{\theta}$ are then chosen such that samples drawn from the marginal probability distribution $P\left({\bm{v}}; \bm{\theta}\right)$ approximate samples drawn from the empirical probability distribution of the data $P_{\textrm{data}}\left(\bm{v}\right)$. This is achieved by minimizing the Kullback--Leibler divergence (KL) divergence~\cite{kullback1951} between $P_{\textrm{data}}\left(\bm{v}\right)$ and $P\left({\bm{v}}; \bm{\theta}\right)$, or equivalently, by minimizing the negative log-likelihood
\begin{equation}
\mathcal{L}\left(\bm{\theta}\right)=-\sum\limits_{\bm{v}}P_{\textrm{data}}\left(\bm{v}\right)\log\left(P\left(\bm{v}; \bm{\theta}\right)\right).
\end{equation}
The minimization of $\mathcal{L}\left(\bm{\theta}\right)$ is usually performed using a gradient based optimizer, where the gradient of the loss function is:
\begin{equation}
\partial_\theta\mathcal{L}\left(\bm{\theta}\right)=\sum\limits_{\bm{v}}P_{\textrm{data}}\left(\bm{v}\right)\expval{\partial_\theta E\left(\bm{v}, \bm{h}; \bm{\theta}\right)}_{\bm{v}}-\expval{\partial_\theta E\left(\bm{v}, \bm{h}; \bm{\theta}\right)}.\label{log-likelihood}
\end{equation}
Here, $\expval{\cdot}$ denotes the average with respect to a Boltzmann distribution with the energy given by Eq.~\eqref{rbmenergyrel}, and $\expval{\cdot}_{\bm{v}}$ is the same but with the visible nodes clamped to $\bm{v}$. The first term is known as the \textit{positive phase}, and the second term as the \textit{negative phase}. Exact maximum likelihood learning of the negative phase requires knowledge of the partition function, which is intractable. Therefore, in practice, this phase is approximated using Gibbs sampling~\cite{4767596}. We use Persistent Contrastive Divergence (PCD)~\cite{Tieleman:2008:TRB:1390156.1390290} to train our RBM and simulated annealing to sample from it; both are described in Appendix~\ref{sec:rbm-train-sampl}.

\subsection{Generative Adversarial Networks}\label{sec:gener-advers-netw}

\textit{Generative adversarial networks} (GANs) are structured probabilistic models over the space of observed variables $\bm{x}$ and latent variables $\bm{z}$~\cite{NIPS2014_5423,goodfellow2016nips}. They implicitly model high-dimensional distributions of data and can be used to efficiently generate samples from these distributions. GANs are generally characterized by a pair of neural networks competing against each other in a zero-sum game. The generator network $\mathcal{G}$ with parameters $\bm{\theta}^{\left(\mathcal{G}\right)}$ learns to map vectors $\bm{z}$ from the latent space to samples $\bm{x} = \mathcal{G}\left(\bm{z}; \bm{\theta}^{\left(\mathcal{G}\right)}\right)$ drawn from a probability distribution that is close to $P_{\textrm{data}}$. The discriminator network $\mathcal{D}$ with parameters $\bm{\theta}^{\left(\mathcal{D}\right)}$ receives samples from both the generator and the true distribution of data, and emits a probability $\mathcal{D}\left(\bm{x}; \bm{\theta}^{\left(\mathcal{G}\right)}\right)\in\left[0,1\right]$ that the input is real. The generator wishes to minimize its loss function $J^{\left(\mathcal{G}\right)}\left(\bm{\theta}^{\left(\mathcal{G}\right)}, \bm{\theta}^{\left(\mathcal{D}\right)}\right)$ by attempting to fool the discriminator into believing that its generated samples are real. The discriminator, on the other hand, tries to minimize its loss function $J^{\left(\mathcal{D}\right)}\left(\bm{\theta}^{\left(\mathcal{G}\right)}, \bm{\theta}^{\left(\mathcal{D}\right)}\right)$ by correctly classifying the inputs as either fake or real~\cite{goodfellow2016deep}. We can cast these statements into the minimax optimization problem~\cite{goodfellow2016nips}:
\begin{equation}\label{ganloss}
\begin{split}
J^{\left(\mathcal{G}\right)}&(\bm{\theta}^{\left(\mathcal{G}\right)}, \bm{\theta}^{\left(\mathcal{D}\right)}) = -J^{\left(\mathcal{D}\right)}(\bm{\theta}^{\left(\mathcal{G}\right)}, \bm{\theta}^{\left(\mathcal{D}\right)}) \\
&= \expval{\log\left(\mathcal{D}\left(\bm{x}\right)\right)}_{\textrm{data}} + \expval{\log\left(1 - \mathcal{D}\left(\mathcal{G}\left(\bm{z}\right)\right)\right)}_{\textrm{noise}},
\end{split}
\end{equation}
where the goal is to find the optimal generator parameters
\begin{equation}\label{optimaltheta}
\bm{\theta}^{\left(\mathcal{G}\right)*} = \argmin_{\bm{\theta}^{\left(\mathcal{G}\right)}} \max_{\bm{\theta}^{\left(\mathcal{D}\right)}} J^{\left(\mathcal{G}\right)}(\bm{\theta}^{\left(\mathcal{G}\right)}, \bm{\theta}^{\left(\mathcal{D}\right)}).
\end{equation}
Here, $\expval{\cdot}_{\textrm{data}}$ denotes the expectation value when $\bm{x}$ is drawn from real data, and $\expval{\cdot}_{\textrm{noise}}$ denotes the expectation value when $\bm{z}$ is drawn from the distribution of latent variables. For GANs, the latent variables are normally chosen to be Gaussian or uniform noise~\cite{NIPS2014_5423}. The first term in Eq.~\eqref{ganloss} favors the discriminator outputting $1$ on real data, while the second term favors the discriminator outputting $0$ on generated data. The generator strives to achieve the opposite. The solution to this optimization problem is a point of Nash equilibrium where the generator samples are indistinguishable from the real data and the discriminator predicts $0.5$ on all inputs. 

Unfortunately, finding the Nash equilibrium is quite difficult in practice~\cite{NIPS2016_6125}. In recent years, there have been many proposals aiming to improve the training of GANs. Some of them include considering a non-saturating version of the generator loss function~\cite{goodfellow2016nips}, introducing surrogate objective functions~\cite{metz2016unrolled}, regularizing the discriminator by adding gradient penalty terms~\cite{gulrajani2017improved,kodali2018convergence}, and using a formulation of the training objective based on the Wasserstein-1 distance~\cite{arjovsky2017wasserstein,gulrajani2017improved}. Most of these architectures are extensions of deep convolutional GANs (DCGANs)~\cite{radford2015unsupervised}, which use convolutional neural networks (CNNs) for their generators and discriminators, instead of the fully-connected dense layers proposed initially~\cite{NIPS2014_5423}. Our implementation of a DCGAN is described in Fig.~\ref{fig:dcgan}. In this paper we focus on the AAN architecture~\cite{arici2016associative}, which is an extension of the DCGAN architecture, and the improvements that it brings to training GANs, as described in Section~\ref{sec:assoc-advers-netw}. 

\begin{figure}
\begin{center}
\resizebox{\linewidth}{!}{
\begin{tikzpicture}
\draw[fill opacity=0.6,rounded corners,text opacity=1,fill=yellow!50,text=black] (-2,0) rectangle node[rotate=90] {Noise} (-1,2);
\draw[->] (-1,1) -- (0,1);
\draw[fill opacity=0.6,fill=green!50,rounded corners,text=green!50!black,text opacity=1] (0,0) rectangle node {$\mathcal{G}$} (2,2);
\draw[->] (2,1) -- (2.5,1);
\draw[fill opacity=0.6,fill=orange!50,text=black,text opacity=1] (3,1) circle (0.5) node {or};
\draw[->] (3,-0.5) -- (3,0.5);
\draw[fill opacity=0.6,rounded corners,text opacity=1,fill=yellow!50,text=black] (2,-1.5) rectangle node {Data} (4,-0.5);
\draw[->] (3.5,1) -- (4,1);
\draw[fill=blue!50,fill opacity=0.6,rounded corners,text=blue,text opacity=1] (4,0) rectangle node {$\mathcal{D}$} (6,2);
\draw[->] (6,1) -- (7,1);
\node[right] at (7,1) {$\left[0,1\right]$};
\end{tikzpicture}
}
\caption{\label{fig:gan}Diagrammatic representation of a GAN. The generator $\mathcal{G}$ and the discriminator $\mathcal{D}$ are trained jointly. The generator learns to forge data, and the discriminator learns to distinguish between forged data and real data.}
\end{center}
\end{figure}

\begin{figure}
\begin{center}
\begin{subfloat}[Deep convolutional generator]{
\resizebox{0.45\linewidth}{!}{
\begin{tikzpicture}
\draw[fill opacity=0.6,text opacity=1,fill=yellow!50,text=black] (-1.5, 0) rectangle node {Noise} (1.5, 0.5);
\draw[->] (0, 0.5) -- (0, 1);
\draw[fill opacity=0.6,text opacity=1,fill=yellow!50,text=black] (-1.5, 1) rectangle node {Fully-connected} (1.5, 1.5);
\draw[fill opacity=0.6,text opacity=1,fill=pink!50,text=black] (-1.5, 1.5) rectangle node {Batch Normalization} (1.5, 2);
\draw[fill opacity=0.6,text opacity=1,fill=cyan!50,text=black] (-1.5, 2) rectangle node {Leaky ReLU} (1.5, 2.5);
\draw[->] (0, 2.5) -- (0, 3);
\draw[fill opacity=0.6,text opacity=1,fill=teal!50,text=black] (-1.5, 3) rectangle node {Deconv $5\times5$} (1.5, 3.5);
\draw[fill opacity=0.6,text opacity=1,fill=pink!50,text=black] (-1.5, 3.5) rectangle node {Batch Normalization} (1.5, 4);
\draw[fill opacity=0.6,text opacity=1,fill=cyan!50,text=black] (-1.5, 4) rectangle node {Leaky ReLU} (1.5, 4.5);
\draw[->] (0, 4.5) --  (0, 5);
\draw[dashed,rounded corners] (-2.5, 2.75) rectangle (2.5, 4.75) {};
\node[left] at (3.25, 3.75) {$\times 3$};
\draw[fill opacity=0.6,text opacity=1,fill=teal!50,text=black] (-1.5, 5) rectangle node {Conv $5\times5$} (1.5, 5.5);
\draw[fill opacity=0.6,text opacity=1,fill=cyan!50,text=black] (-1.5, 5.5) rectangle node {$\tanh$} (1.5, 6);
\draw[->] (0, 6) -- (0, 6.5);
\draw[fill opacity=0.6,text opacity=1,fill=yellow!50,text=black] (-1.5, 6.5) rectangle node {Data} (1.5, 7);
\end{tikzpicture}\label{fig:gen}
}
}
\end{subfloat}
\begin{subfloat}[Deep convolutional discriminator]{
\resizebox{0.45\linewidth}{!}{
\begin{tikzpicture}
\draw[fill opacity=0.6,text opacity=1,fill=yellow!50,text=black] (-1.5, 0) rectangle node {Data} (1.5, 0.5);
\draw[->] (0, 0.5) -- (0, 1);
\draw[fill opacity=0.6,text opacity=1,fill=teal!50,text=black] (-1.5, 1) rectangle node {Conv $5\times5$} (1.5, 1.5);
\draw[fill opacity=0.6,text opacity=1,fill=pink!50,text=black] (-1.5, 1.5) rectangle node {Batch Normalization} (1.5, 2);
\draw[fill opacity=0.6,text opacity=1,fill=cyan!50,text=black] (-1.5, 2) rectangle node {Leaky ReLU} (1.5, 2.5);
\draw[->] (0, 2.5) -- (0, 3);
\draw[dashed,rounded corners] (-2.5, 0.75) rectangle (2.5, 2.75) {};
\node[left] at (3.25, 1.75) {$\times 4$};
\draw[fill opacity=0.6,text opacity=1,fill=yellow!50,text=black] (-1.5, 3) rectangle node {Fully-connected} (1.5, 3.5);
\draw[fill opacity=0.6,text opacity=1,fill=green!50,text=black] (-1.5, 3.5) rectangle node {Clamped} (1.5, 4);
\draw[fill opacity=0.6,text opacity=1,fill=yellow!50,text=black] (-1.5, 4) rectangle node {Fully-connected} (1.5, 4.5);
\draw[fill opacity=0.6,text opacity=1,fill=cyan!50,text=black] (-1.5, 4.5) rectangle node {$\operatorname{\upsigma}$} (1.5, 5);
\draw[->] (0, 5) -- (0, 5.5);
\draw[fill opacity=0.6,text opacity=1,fill=yellow!50,text=black] (-1.5, 5.5) rectangle node {Prediction} (1.5, 6);
\end{tikzpicture}\label{fig:disc}
}
}
\end{subfloat}
\caption{\label{fig:dcgan}Architecture layout of the deep convolutional neural networks composing our DCGAN. Although not used in DCGANs, the green block denotes clamping to a Boltzmann machine (either classical or quantum) as used in the AAN and QAAN architectures.}
\end{center}
\end{figure}

\subsection{Associative Adversarial Networks}\label{sec:assoc-advers-netw}

An associative memory provided by an RBM can circumvent the imbalance in training rates typically present in a GAN and enhance the expressivity of the model; such architectures are called \textit{associative adversarial networks} (AANs)~\cite{arici2016associative}. In an AAN, the latent space for the generator is treated as a feature space that is learned by the RBM. The RBM is simultaneously trained with the GAN on an intermediate layer of the discriminator, with a sigmoid activation function interpreted as the probability of a particular neuron firing. In general, in an AAN, the discriminator $\mathcal{D}$ is decomposed into a mapping into the feature space $\mathcal{F}$ and a a classifier $\mathcal{C}$ such that:
\begin{equation}
\mathcal{D}=\mathcal{C}\circ\mathcal{F}.
\end{equation}
We find that in practice, however, using a trivial classifier as in Fig.~\ref{fig:disc} suffices to improve the performance of the generator.

\begin{figure}
\begin{center}
\resizebox{\linewidth}{!}{
\begin{tikzpicture}
\draw[fill opacity=0.6,rounded corners,text opacity=1,fill=yellow!50,text=black] (-2,0) rectangle node[rotate=90] {\small Samples} (-1,2);
\draw[->] (-1,1) -- (0,1);
\draw[fill opacity=0.6,fill=green!50,rounded corners,text=green!50!black,text opacity=1] (0,0) rectangle node {$\mathcal{G}$} (2,2);
\draw[->] (2,1) -- (2.5,1);
\draw[fill opacity=0.6,fill=orange!50,text=black,text opacity=1] (3,1) circle (0.5) node {or};
\draw[->] (3,-0.5) -- (3,0.5);
\draw[fill opacity=0.6,rounded corners,text opacity=1,fill=yellow!50,text=black] (2,-1.5) rectangle node {Data} (4,-0.5);
\draw[->] (3.5,1) -- (4,1);
\draw[fill=blue!50,fill opacity=0.6,rounded corners,text=blue,text opacity=1] (4,0) rectangle node {$\mathcal{D}$} (6,2);
\draw[fill=cyan,dashed] (5.7,0) rectangle (5.8,2);
\draw[cyan!50!black] (5.75,0) to[out=270,in=0] node[below,black,rotate=60] {Clamped} (4,-3);
\draw[cyan!50!magenta] (4,-3) -- (3,-3);
\draw[cyan!50!magenta] (3,-3) -- (2,-3);
\draw[->,magenta!50!black] (2,-3) to[out=180,in=270] node[below,black,rotate=-45] {Sampled} (-1.5,0);
\draw[->] (6,1) -- (7,1);
\node[right] at (7,1) {$\left[0,1\right]$};
\draw (2.5, -4) -- (2, -3);
\draw (2.5, -4) -- (3, -3);
\draw (2.5, -4) -- (4, -3);
\draw (3.5, -4) -- (2, -3);
\draw (3.5, -4) -- (3, -3);
\draw (3.5, -4) -- (4, -3);
\node[draw,circle,fill=blue!50,minimum size=15,inner sep=0] at (2.5, -4) {};
\node[draw,circle,fill=blue!50,minimum size=15,inner sep=0] at (3.5, -4) {};
\node[draw,circle,fill=green!50,minimum size=15,inner sep=0] at (2, -3) {};
\node[draw,circle,fill=green!50,minimum size=15,inner sep=0] at (3, -3) {};
\node[draw,circle,fill=green!50,minimum size=15,inner sep=0] at (4, -3) {};
% \tikzstyle{neuron}=[circle,minimum size=0.5cm]
% \tikzstyle{input neuron}=[neuron, fill=green!50];
% \tikzstyle{hidden neuron}=[neuron, fill=blue!50];
% \foreach\name / \y in {1,...,4}
% \node[input neuron] (V-\name) at (5.5-\y,-3) {};
% \foreach\name / \y in {1,...,3}
% \path[xshift=-0.5cm]
% node[hidden neuron] (H-\name) at (5.5-\y,-4) {};
% \foreach\source in {1,...,4}
% \foreach\dest in {1,...,3}
% \draw (V-\source) -- (H-\dest);
\end{tikzpicture}
}
\caption{\label{fig:aan}Schematic representation of an AAN. The RBM acts as an associative memory for the GAN. The RBM is simultaneously trained with the GAN to learn the distribution of a layer of the discriminator of the GAN. The distribution approximated by the RBM then acts as the latent distribution for the generator of the GAN (instead of noise). A QAAN has an identical architecture, except that the RBM is replaced with a QBM.}
\end{center}
\end{figure}
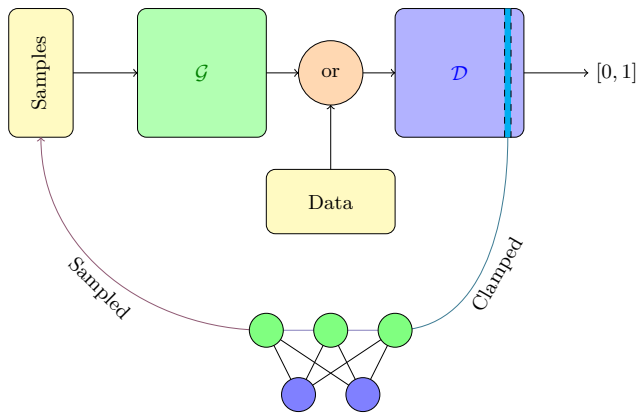

Though the initial motivation behind AANs was to bring the generator and discriminator learning rates in line with each other---as often the instability in GAN training is due to the discriminator learning more quickly than the generator---our DCGAN implementation for the data sets under consideration experiences no notable difference in the learning rates of the generator and discriminator, and thus our main advantage when using an AAN is due to the generator expressivity gained when it draws samples from an improved feature space rather than from noise. Our AAN architecture consists of the DCGAN architecture described in Fig.~\ref{fig:dcgan}, coupled to an RBM as in Fig.~\ref{fig:aan}.

\section{Quantum Generative Models}\label{sec:near-term-quantum}

\subsection{Quantum Boltzmann Machines}\label{sec:quant-boltzm-mach}

\textit{Quantum Boltzmann machines} (QBMs) are a recently introduced method of quantizing Boltzmann machines that have been numerically observed to give a quantum speedup in both the rate of training and in the accuracy of the approximating distributions~\cite{amin2016quantum,PhysRevA.96.062327}. As initially proposed, instead of considering the classical energy function of Eq.~\eqref{bmerel}, one considers the Hamiltonian:
\begin{equation}\label{qbmhamiltonianrel}
H\left(\bm{\theta}\right)=-\sum\limits_a\varGamma_a\sigma_a^x-\sum\limits_a b_a\sigma_a^z-\sum\limits_{a<b}W_{ab}\sigma_a^z\sigma_b^z
\end{equation}
and the thermal density matrix:
\begin{equation}\label{thermalrhorel}
\rho\left(\bm{\theta}\right)=\frac{\ce^{-H\left(\bm{\theta}\right)}}{\tr\left(\ce^{-H\left(\bm{\theta}\right)}\right)},
\end{equation}
where now $\bm{\theta}=\left\{\bm{\varGamma},\bm{b},\bm{W}\right\}$ are the model parameters. Defining the projector onto the subspace with visible nodes equal to $\bm{v}$ as $\varPi_{\bm{v}}$, our goal is to now train the parameters $\bm{\varGamma}$, $\bm{b}$, and $\bm{W}$ such that the probability distribution of samples from $\rho$
\begin{equation}
P\left(\bm{v};\bm{\theta}\right)=\tr\left(\varPi_{\bm{v}}\rho\left(\bm{\theta}\right)\right)
\end{equation}
approximates the empirical probability distribution $P_{\textrm{data}}\left(\bm{v}\right)$. Due to the difficulties in computing the log-liklihood of this distribution~\cite{amin2016quantum}, one instead usually~\footnote{The recently introduced relative entropy training~\cite{PhysRevA.96.062327,2019arXiv190509902W} is approximation free, but we do not study its behavior in this work.} trains QBMs to minimize the upper bound on the loss function
\begin{equation}\label{tildemathcallrel}
\mathcal{L}\left(\bm{\theta}\right)\leq\tilde{\mathcal{L}}\left(\bm{\theta}\right)\equiv-\sum\limits_{\bm{v}}P_{\textrm{data}}\left(\bm{v}\right)\left(\frac{\tr\left(\ce^{-H_{\bm{v}}}\left(\bm{\theta}\right)\right)}{\tr\left(\ce^{-H\left(\bm{\theta}\right)}\right)}\right),
\end{equation}
where
\begin{equation}
H_{\bm{v}}\left(\bm{\theta}\right)=H\left(\bm{\theta}\right)-\ln\left(\varPi_{\bm{v}}\right)=H\left(\sigma_\upsilon^x\to0,\sigma_\upsilon^z\to v_\upsilon;\bm{\theta}\right)
\end{equation}
is the \textit{clamped Hamiltonian}. The gradients of this loss function are now given by:
\begin{equation}\label{qbmgradrel}
\partial_\theta\tilde{\mathcal{L}}\left(\bm{\theta}\right)=\sum\limits_{\bm{v}}P_{\textrm{data}}\left(\bm{v}\right)\expval{\partial_\theta H_{\bm{v}}\left(\bm{\theta}\right)}_{\bm{v}}-\expval{\partial_\theta H\left(\bm{\theta}\right)},
\end{equation}
where here $\expval{\cdot}$ denotes expectation values taken with respect to $\rho$ and $\expval{\cdot}_{\bm{v}}$ denotes expectation values taken with respect to the thermal density matrix with clamped Hamiltonian $H_{\bm{v}}$. Training on this upper bound leads to $\bm{\varGamma}\to\bm{0}$, so in training $\bm{\varGamma}$ is fixed to some constant and treated as a learning hyperparameter~\cite{amin2016quantum}.

In order to numerically simulate large QBMs on a classical computer, we consider only the stoquastic Hamiltonian given by Eq.~\eqref{qbmhamiltonianrel}, and not more general \textsf{QMA}-hard Hamiltonians which have been studied in the context of QBMs~\cite{PhysRevA.96.062327,2019arXiv190205162W}. The details of our Monte Carlo-based simulation method are given in Appendix~\ref{sec:qbm-train-sampl}.

\subsection{Quantum-Classical Associative Adversarial Networks}\label{sec:quant-assoc-advers}

We now quantize AANs by transforming the associated RBM into a QBM, and call the resulting architecture a \textit{quantum-classical associative adversarial network} (QAAN). Our implementation otherwise exactly follows that of our AAN (see Sec.~\ref{sec:assoc-advers-netw}).

The hybrid quantum-classical nature of our QAAN architecture lends itself well to being implementable on NISQ devices. Though in general simulating quantum thermal states is inefficient even on quantum devices, there is evidence that the structure of QBMs allows for efficient heuristic training on NISQ devices~\cite{2019arXiv190301359A}. Furthermore, as QBM training only necessitates measuring simple local observables in the QBM state, there is no need for a quantum memory for the training data. Finally, as the QBM is only trained on a feature space of much lower dimensionality than the data, many fewer qubits are required to implement the QBM than if it were to directly learn the data. This hybrid quantum-classical approach to machine learning, inspired by variational hybrid quantum-classical algorithms~\cite{farhi2014quantum,mcclean2016theory,peruzzo2014variational}, can serve as a model for similar future quantum machine learning architectures that minimize the need for large quantum devices.

\section{Results}\label{sec:numerical-results}

We compare the performance of the classical and quantum-classical architectures on three data sets of increasing difficulty. First, we show that QBMs outperform RBMs on a simple synthetic data set of mixed Bernoulli distributions, thus suggesting that quantum models can provide an improvement in approximating certain distributions. Next, we compare the performance of DCGAN, AAN, and QAAN architectures on the MNIST data set~\cite{lecun1998mnist}, which is a standard benchmark used in classical machine learning. Finally, we test the three architectures on the more challenging CIFAR-10 data set~\cite{krizhevsky2009learning}, and show that our implementation of a QAAN architecture produces samples that more closely mirror the samples drawn from the data distributions.

\begin{figure}
\begin{center}
\includegraphics[width=\columnwidth]{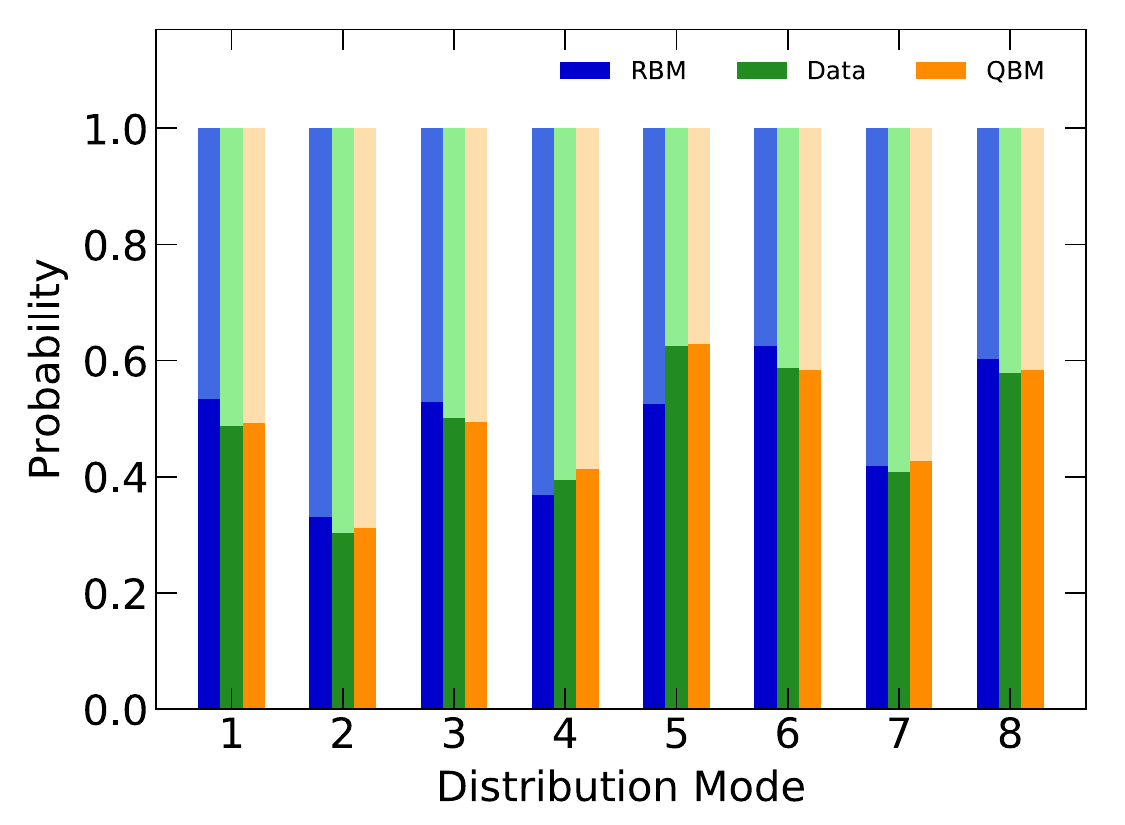}
\caption{\label{fig:rbmqbmcomp}The results of training an RBM and a QBM on a data distribution given by a mixture of multidimensional Bernoulli distributions (see Sec.~\ref{sec:synthetic-data}). The dark regions denote the probability of a Bernoulli variable to be $1$, and the light regions denote the probability of a Bernoulli variable to be $0$. The QBM distribution matches data more closely than the RBM distribution, which is reflected in its lower KL divergence with the data distribution.}
\end{center}
\end{figure}

In order to quantitatively evaluate the different architectures on real data sets, we use the \textit{Inception score}~\cite{NIPS2016_6125} and the \textit{Fr\'{e}chet Inception distance} (FID)~\cite{heusel2017}. The Inception score computes the KL divergence between the conditional class distribution and the marginal class distribution over generated samples as predicted by an Inception-v3 network~\cite{szegedy2016rethinking}. A higher score indicates better generated images. This is the most widely used metric for evaluating GANs and allows for easy comparisons with previous works. However, the IS has some limitations in assessing the realism and intra-class diversity of the generated samples. The FID is a more comprehensive metric that has been shown to mitigate some of these shortcomings~\cite{heusel2017}. It relies on computing the Wasserstein-2 distance between the generated data and the real data in the feature space of
an Inception-v3 network. Lower FID values are better and suggest that the generated images are more similar to the original data set. A detailed description of both metrics is provided in Appendix~\ref{sec:metrics}. 

\begin{figure}[ht]
\begin{center}
\includegraphics[width=\columnwidth]{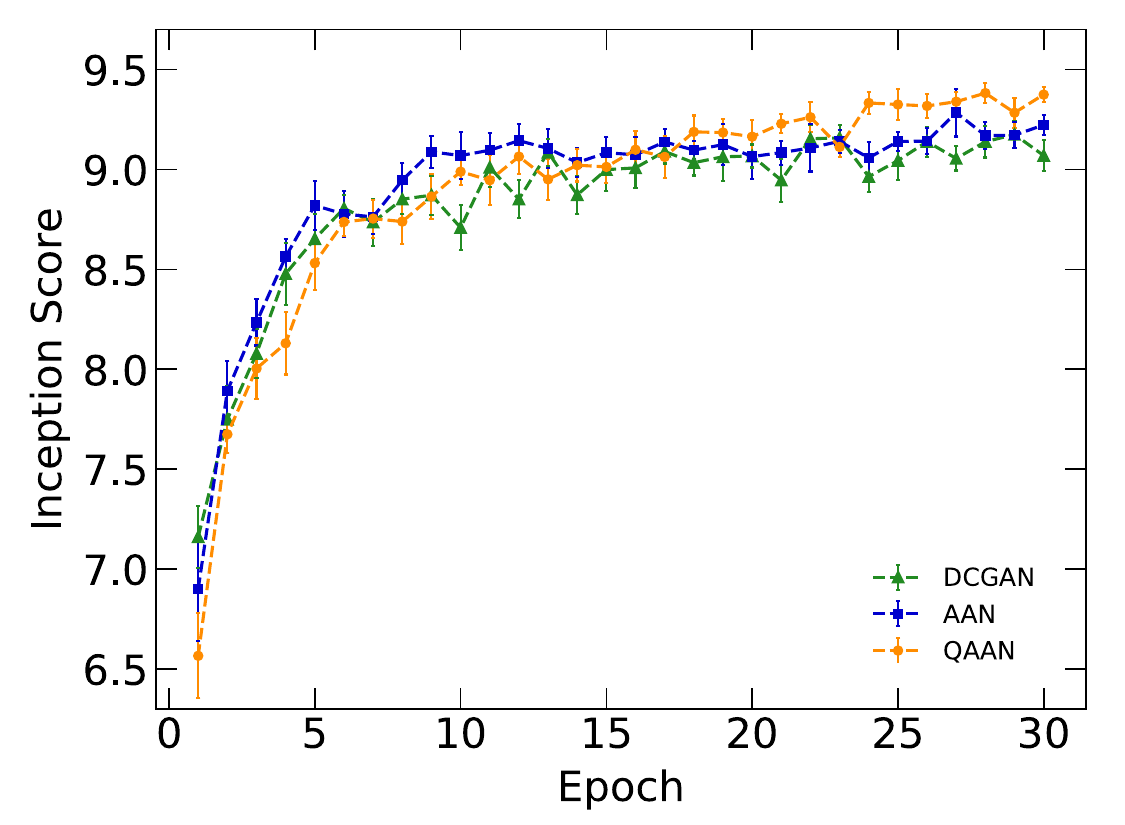}
\caption{\label{fig:inceptionresults}The Inception score of our three generative models trained on the MNIST data set as a function of training epoch. The QAAN clearly outperforms both classical architectures. Error bars denote the standard deviation of the mean Inception score. The MNIST test set has an Inception score of approximately $9.75$.}
\end{center}
\end{figure}

\begin{figure*}[ht]
\begin{center}
\includegraphics[width=\textwidth]{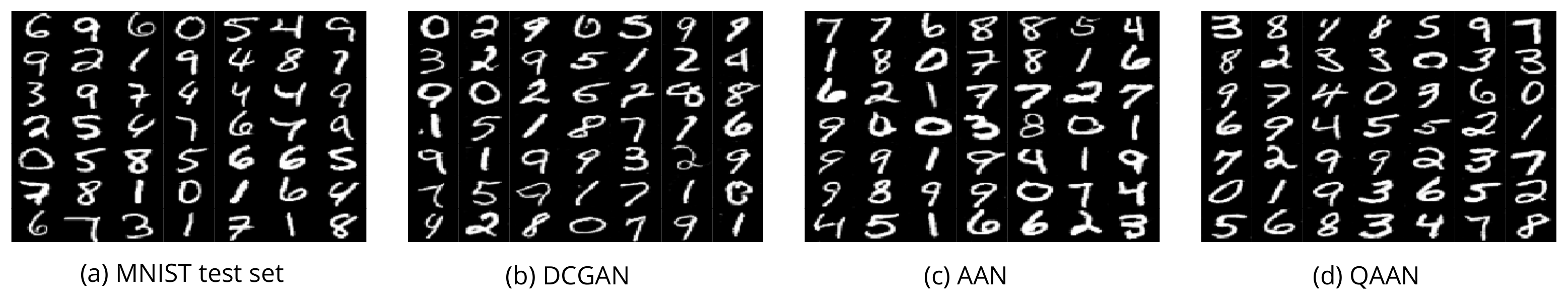}
\caption{\label{fig:genmnist}Samples of handwritten digits from the original MNIST data set and from our generative models.}
\end{center}
\end{figure*}

In all of our experiments, $10^4$ samples are randomly drawn from each model and used to evaluate its performance. Similar results are obtained for different initializations of the network parameters. The Inception score showed a higher variance over different data subsets and therefore we report the average over $10$ batches, each consisting of $1000$ generated images. Since the FID relies on a particular layer of a pre-trained network and can only be unambigously defined for colored images, we only use it in quantifying the performace of our models on CIFAR-10. 

\subsection{Synthetic Data}\label{sec:synthetic-data} 

We begin by comparing the learning capability of an RBM with that of a QBM by training both on $6400$ samples from the mixed Bernoulli distribution:
\begin{equation}
P_{\textrm{data}}\left(\bm{z};\left\{\bm{s}^i\right\}\right)=\sum\limits_{i=1}^N q^{d\left(\bm{z},\bm{s}^i\right)}\left(1-q\right)^{\dim{\left(\bm{z}\right)}-d\left(\bm{z},\bm{s}^i\right)}
\end{equation}
with $N$ random modes $\left\{\bm{s}^i\right\}$, where $d\left(\bm{a},\bm{b}\right)$ denotes the Hamming distance between $\bm{a}$ and $\bm{b}$. We choose $N=8$ and $q=0.9$ for our numerical experiments. More details about our training procedure are provided in Appendix~\ref{sec:training-parameters}. Samples are then drawn from each of the data, RBM, and QBM distributions. The resulting empirical probability distributions are given by Fig.~\ref{fig:rbmqbmcomp}. To quantitatively measure the distance between the original and reconstructed distributions, we compute the KL divergence $\KLD{P_{\textrm{data}}}{P_{\textrm{BM}}}$. We obtain a KL divergence of approximately $1.23$ for the RBM and $0.76$ for the QBM. We see that the QBM outperforms the RBM, even though both of them have the same number of parameters and are trained similarly.

\subsection{MNIST Data}\label{sec:mnist-data}

We now compare our implementations of DCGAN, AAN, and QAAN. We train all three networks on the MNIST handwritten digit data set~\cite{lecun1998mnist}, which consists of $6\times 10^4$ grayscale images, each of size $28\times 28$ pixels. We rescale the pixel values to the interval $\left[-1, 1\right]$ before feeding them into our networks. The training parameters for this data set are summarized in Appendix~\ref{sec:training-parameters}.

We monitor the performance of each network by computing the Inception score on generated images after every epoch, as shown in Fig.~\ref{fig:inceptionresults}. We notice that the results converge within the considered $30$ epochs, with QAAN reaching a better Inception score than its classical counterpart by roughly $2\%$. It is worth mentioning that all of the architectures considered achieve a score that is close to that of real data ($9.75$). We also visually examine samples of generated handwritten digits, as shown in Fig.~\ref{fig:genmnist}, and confirm that they are almost indistinguishable from the orginal data.

\begin{figure}
\begin{center}
\includegraphics[width=\columnwidth]{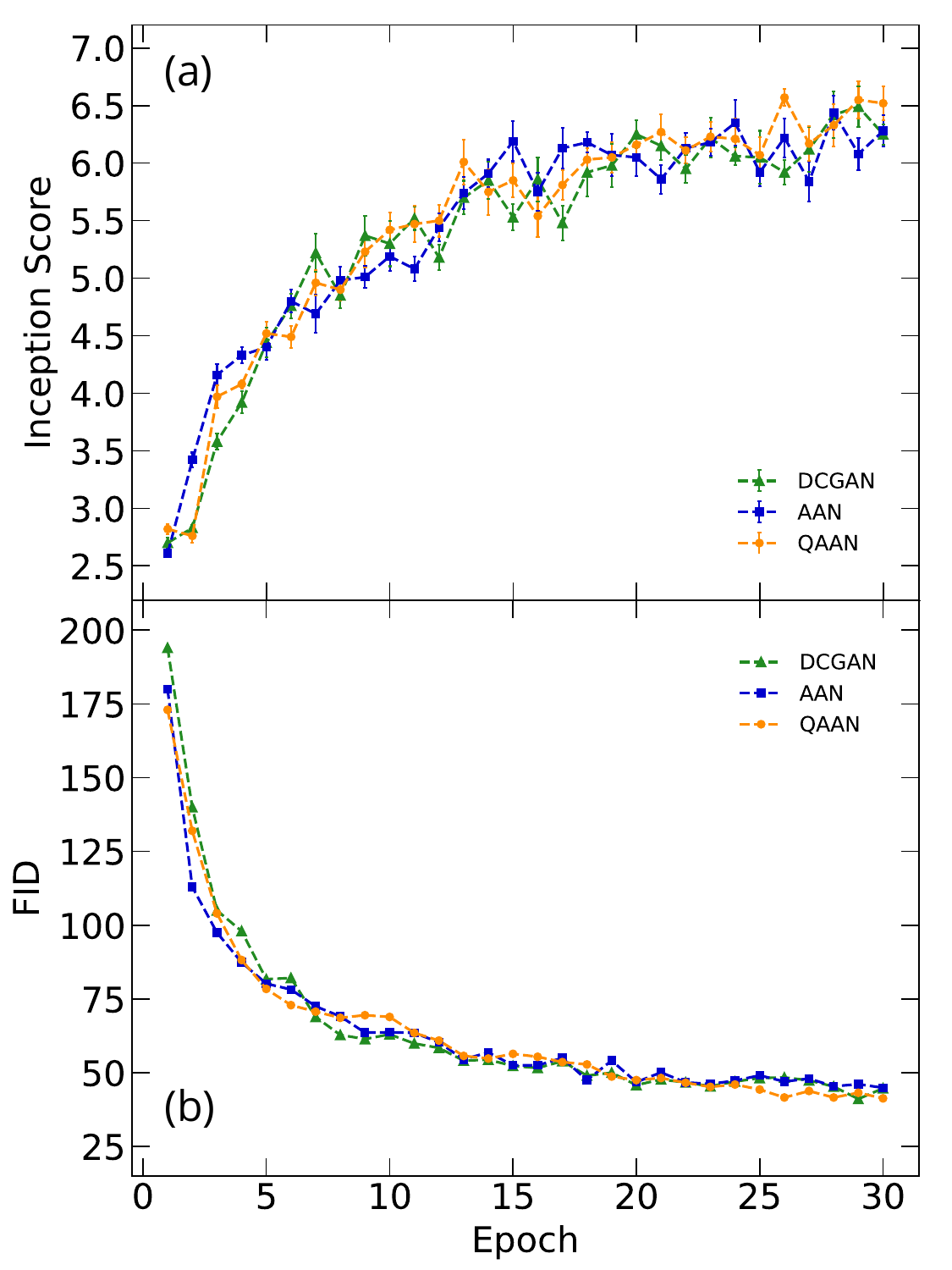}
\caption{\label{fig:cifarresults}Evaluation of our generative models trained on the CIFAR-10 data set using (a) the Inception score and (b) the FID as a function of training epoch. In both metrics, the QAAN slightly outperforms both the DCGAN and the AAN. Error bars denote the standard deviation of the mean Inception score. The CIFAR-10 test set has an Inception score of $11.31$ and an FID score of $3.16$.}
\end{center}
\end{figure}

\begin{figure*}
\begin{center}
\includegraphics[width=\textwidth]{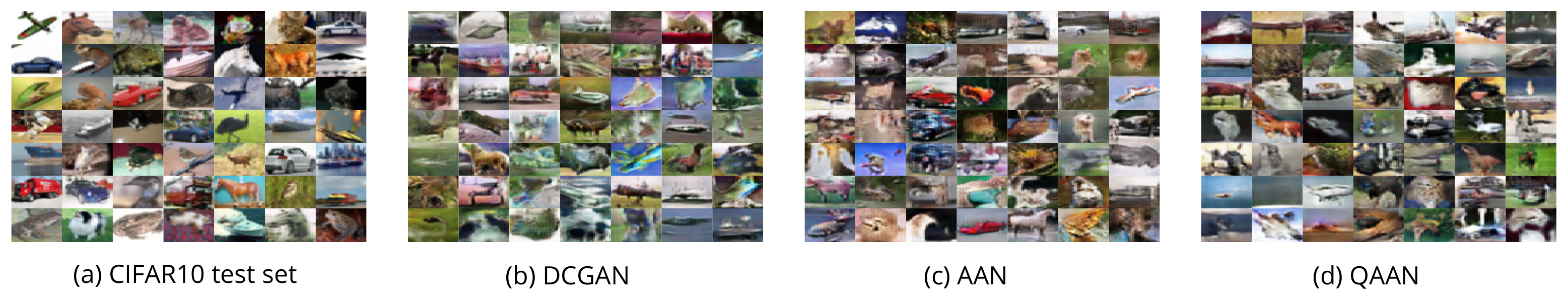}
\caption{\label{fig:gencifar}Samples of CIFAR-10 images from the original data set and from our generative models.}
\end{center}
\end{figure*}

\subsection{CIFAR-10 Data}
\label{sec:CIFAR-10}

Finally, we study the performance of our models on the CIFAR-10 data set. This data set consists of $6\times 10^4$ colored natural scene images, each of size $32\times 32$ pixels and $3$ color channels, divided across $10$ different classes~\cite{krizhevsky2009learning}. The training procedure is identical to the case of MNIST and is described in Appendix~\ref{sec:training-parameters}.

We once again compute the Inception score and FID on generated images after every epoch, and plot our results for all three models in Fig.~\ref{fig:cifarresults}. Notice that the inception scores are overall lower than in the case of MNIST, even though both data sets have $10$ classes. This can be attributed to the increased complexity of the data set of colored images. The better performance of QAAN is more prominent in the FID metric, where it achieves a consistently better score in the last five training epochs when compared to DCGAN and AAN. The improvement in FID is approximately $7\%$ by the end of training. The metrics reported in Fig.~\ref{fig:cifarresults} are on par with those obtained for a variety of classical GANs, such as WGANs, with similar architectures~\cite{fedus2017many,ostrovski2018autoregressive}. We also note that both AAN and QAAN have a steeper learning curve in the early epochs, which can be attributed to the associative memory providing a more meaningful latent space for the generator. Sample images from CIFAR-10 and our models are shown in Fig.~\ref{fig:gencifar}. The generated images look very realistic.

\section{Discussion}\label{sec:discussion}

In Sec.~\ref{sec:numerical-results} we showed that the QAAN architecture can learn the MNIST and CIFAR-10 data sets more effectively than the AAN and DCGAN architectures. Since the only difference between the QAAN and AAN architectures is the use of a QBM, rather than an RBM, we attribute to it the observed increase in performance. Nonetheless, QAAN's learning advantage is not as substantial as QBM's edge on simpler data sets, as supported by~\cite{amin2016quantum,PhysRevA.96.062327} and our results in Sec.~\ref{sec:synthetic-data}. We suspect that this is due to the moderate size of the QBM used in our QAAN architecture. The bottleneck in our numerical experiments comes from simulating the QBM qubits on a classical computer through Monte Carlo sampling, which severely limits the accessible system sizes for the QAAN associative memory. Further improvements may be gained by improving our classical simulations and by considering \textsf{QMA}-hard Hamiltonians~\cite{PhysRevA.96.062327,2019arXiv190205162W}, which we leave to future work.

As QAANs are a quantum-assisted classical architecture, they lend themselves well to potential experimental implementations on NISQ devices. There are proposals for implementing QBMs on quantum annealing devices~\cite{PhysRevA.92.052323} and generic NISQ devices~\cite{2019arXiv190301359A}, and the necessary Gibbs distributions have been produced on atomic lattice systems with similar Hamiltonians through Hamiltonian quenching~\cite{bernien2017probing,Kaufman794}. Furthermore, as the necessary number of visible units of the QBM only grows as the dimensionality of the latent space of the QAAN---which in general is much smaller than the size of the probability distribution being approximated---the necessary number of qubits also remains small. For instance, our simulations considered Boltzmann machines with $32$ visible units and $8$ hidden units (see Appendix~\ref{sec:training-parameters}), whereas the dimensionality of the MNIST input data is $784$ and the dimensionality of the CIFAR-10 input data is $3072$. These considerations suggest that a QAAN could be implemented in the very near future.

While preparing this manuscript, we became aware of a similar project that also considered quantum-assisted classical AANs~\cite{2019arXiv190410573W}. We believe that our work, although similar in network architecture and scope, is still very different when it comes to model training and testing. In particular, we use quantum Monte Carlo to train the QBM, while the authors in~\cite{2019arXiv190410573W} had access to a quantum annealing platform with many more qubits than our simulations could achieve. Nonetheless, our quantum-classical implementation seems to yield better generative performance under the considered metrics. Furthermore, we test our models on a more complex data set of colored images.

In conclusion, we have introduced a new hybrid quantum-classical generative model capable of successfully learning distributions over complex data sets. We have showed numerically that our model slightly outperforms analogous classical generative architectures when trained under similar conditions. In addition, the model could potentially be experimentally tested on NISQ devices. Our work adds to the rapidly-expanding family of quantum-enhanced machine learning algorithms. 

\begin{acknowledgments}
We are grateful to Maxwell Nye for insightful discussion on classical machine learning. We would also like to thank Aram Harrow for his guidance on this project. ERA is partially supported by a Francis C.\ W.\ Greenlaw fellowship, the Henry W.\ Kendall Fellowship Fund, and a Lester Wolfe fellowship from MIT. CZ is partially supported by a Whiteman fellowship from MIT.
\end{acknowledgments}

\bibliography{QAAN}

\appendix

\section{RBM Training and Sampling}\label{sec:rbm-train-sampl}

We can explicitly write the gradients for each of the RBM parameters by substituting Eq.~\eqref{rbmenergyrel} into Eq.~\eqref{log-likelihood}:
\begin{align}
\partial_{b_\upsilon}\mathcal{L}\left(\bm{\theta}\right) &=\sum\limits_{\bm{v}}P_{\textrm{data}}\left(\bm{v}\right)\expval{v_\upsilon}_{\bm{v}}-\expval{v_\upsilon},\\
\partial_{b_\eta}\mathcal{L}\left(\bm{\theta}\right) &=\sum\limits_{\bm{v}}P_{\textrm{data}}\left(\bm{v}\right)\expval{h_\eta}_{\bm{v}}-\expval{h_\eta},\\
\partial_{W_{\upsilon\eta}}\mathcal{L}\left(\bm{\theta}\right) &=\sum\limits_{\bm{v}}P_{\textrm{data}}\left(\bm{v}\right)\expval{v_\upsilon h_\eta}_{\bm{v}}-\expval{v_\upsilon h_\eta}.
\end{align}
Although the probability distribution over the visible and hidden units $P\left({\bm{v}}, {\bm{h}}; \bm{\theta}\right)$ is generally intractable for computing expectation values, the bipartite structure of the RBM makes it very simple to heuristically sample from the conditional probability distributions:
\begin{align}
P\left(h_\eta = 1 \given {\bm{v}}\right)&=\mathop{\upsigma}\left(\sum_{\upsilon}v_\upsilon W_{\upsilon\eta} + b_\eta\right),\label{gibbs}\\
P\left(v_\upsilon = 1 \given {\bm{h}}\right)&=\mathop{\upsigma}\left(\sum_{\eta}W_{\upsilon\eta}h_\eta + b_\upsilon\right),\label{gibbs2}
\end{align}
where $\operatorname{\upsigma}\left(x\right) = \left(1 + \exp\left(-x\right)\right)^{-1}$ is the sigmoid function. We can use these relations to approximate the negative phase of the gradient during training.

Two common algorithms used for this task are Contrastive Divergence (CD)~\cite{hinton2002training} and Persistent Contrastive Divergence (PCD)~\cite{tieleman2008training}. CD uses a data sample $\bm{v}_{\textrm{data}}$ as a starting point of a Markov chain at every optimization iteration and then performs block Gibbs sampling for $k$ steps using Eqs.~\eqref{gibbs} and~\eqref{gibbs2}. PCD instead relies on a single Markov chain with a persistent state that is preserved at the end of each optimization iteration and passed on as the initial state of the next iteration. In practice, choosing $k\in\left[1, 10\right]=O\left(1\right)$ has been shown to be sufficient~\cite{hinton2012practical}. 

In order to generate samples from the RBM, we apply the same procedure, but instead of initializing the visible units with a data sample, we use a random initialization. Furthermore, we use simulated annealing~\cite{kirkpatrick1983optimization} by introducing an inverse temperature $\beta$ that scales the energy function, and hence the model parameters. We initialize our Markov chain with a sample drawn from a uniform distribution at $\beta=0$ and then gradually lower the temperature at each step of the Markov chain until we reach $\beta=1$, which corresponds to the desired distribution parameters. We use a linear schedule in $\beta$ for annealing. This procedure improves the diversity of our samples by increasing the mixing rate of the chain and helps the training procedure avoid getting trapped in local minima of the loss function.

Note that thus far we have assumed that our inputs---given by the visible units---are binary vectors. In the case of real-valued data, such as pixel values of images, we can often rescale the input to be in the range $\left[0, 1\right]$ and treat it as the expectation value $p_i$ of a binary variable~\cite{goodfellow2016deep}. We can then sample each entry $i$ in our visible vector $\bm{v}$ from a Bernoulli distribution with mean $p_i$. It is important to mention that although there are formulations of RBMs with real-valued inputs, known as Gaussian-Bernoulli RBMs, here we use Bernoulli RBMs as they are generally easier to train~\cite{hinton2012practical}.

\section{QBM Training and Sampling}\label{sec:qbm-train-sampl}

We begin by rewriting Eq.~\eqref{qbmgradrel} for each of the QBM parameters, such that at every training step we must estimate:
\begin{align}
-\partial_{b_a}\tilde{\mathcal{L}}\left(\bm{\theta}\right)&=\sum\limits_{\bm{v}}P_{\textrm{data}}\left(\bm{v}\right)\expval{\sigma_a^z}_{\bm{v}}-\expval{\sigma_a^z},\\
-\partial_{W_{ab}}\tilde{\mathcal{L}}\left(\bm{\theta}\right)&=\sum\limits_{\bm{v}}P_{\textrm{data}}\left(\bm{v}\right)\expval{\sigma_a^z\sigma_b^z}_{\bm{v}}-\expval{\sigma_a^z\sigma_b^z}.
\end{align}
As mentioned in Sec.~\eqref{sec:quant-boltzm-mach}, the use of the approximate lower bound given by Eq.~\eqref{tildemathcallrel} precludes the training of $\bm{\varGamma}$. The positive phases of the gradient can be calculated exactly, as given in~\cite{amin2016quantum}:
\begin{equation}
\expval{\sigma_\eta^z}_{\bm{v}}=\frac{b_\eta^{\textrm{eff}}}{D_\eta}\tanh\left(D_\eta\right),
\end{equation}
where
\begin{align}
b_\eta^{\textrm{eff}}&=b_\eta+\sum\limits_\upsilon v_\upsilon W_{\upsilon\eta},\\
D_\eta&=\sqrt{\varGamma_\eta^2+\left(b_\eta^{\textrm{eff}}\right)^2}.
\end{align}

As in the RBM case, it is difficult to estimate the negative phase of the gradient as it requires sampling from a quantum thermal state, a problem which in general is \textsf{NP}-hard~\cite{1982JPhA...15.3241B}. To approximately sample from our numerically simulated QBM, we perform population-annealed Monte Carlo sampling. In the population annealing sampling heuristic~\cite{doi:10.1063/1.1632130}, a population of $R_0$ replicas of the system in question is maintained at infinite temperature, and then cooled to some finite temperature by an annealing schedule of $l$ steps, which is analogous to the number of Gibbs steps $k$ in the training of RBMs. With each cooling step, replicas are duplicated or deleted based on an estimate of their relative Boltzmann weights, and are equilibrated according to some Monte Carlo algorithm~\cite{PhysRevE.82.026704}. By sampling a population from our quantum Boltzmann distribution, we are able to parallelize the generation of a mini-batch of samples with one run of the algorithm.

To perform Monte Carlo sampling at each equilibration step, we use the Trotter--Suzuki mapping of the stoquastic Hamiltonian to a classical energy function with an extra imaginary time dimension, which is discretized into $M$ imaginary time slices~\cite{doi:10.1143/PTP.56.1454}. Under this mapping, the quantum thermal distribution at inverse temperature $\beta$ given by the stoquastic Hamiltonian is approximated by the classical thermal distribution:
\begin{equation}\label{eq:1}
p_\beta\left(\left\{\bm{z}^m\right\}\right)=\frac{\ce^{-\beta\left(E_{\textrm{cl}}\left(\left\{\bm{z}^m\right\}\right)+E_{\textrm{qm}}\left(\left\{\bm{z}^m\right\};\beta\right)\right)}}{\sum\limits_{\left\{\bm{z}^m\right\}}\ce^{-\left(E_{\textrm{cl}}\left(\left\{\bm{z}^m\right\}\right)+E_{\textrm{qm}}\left(\left\{\bm{z}^m\right\};\beta\right)\right)}},
\end{equation}
where
\begin{align}
E_{\textrm{cl}}\left(\left\{\bm{z}^m\right\};\bm{\theta}\right)&=\frac{1}{M}\sum\limits_{m=1}^M E\left(\bm{z}^m;\bm{\theta}\right),\\
E_{\textrm{qm}}\left(\left\{\bm{z}^m\right\};\bm{\theta};\beta\right)&=\frac{1}{2\beta}\sum\limits_{a,m}\ln\left(\tanh\left(\frac{\beta\varGamma_a}{M}\right)\right)z_a^m z_a^{m+1}.
\end{align}
We impose periodic boundary conditions, such that $m=M+1$ is identified with $m=1$. In the limit $M\to\infty$, this approximation is exact. Then, we perform the necessary Monte Carlo sampling using the Metropolis--Hastings algorithm~\cite{10.1093/biomet/57.1.97} on the mapped set of spins.

In our simulations, we use $R_0=64$ initial population replicas corresponding to a single mini-batch, a linear annealing schedule of $l=5$ steps from $\beta=0$ to $\beta=1$, and $M=10$ imaginary time slices.

\section{Training Parameters}\label{sec:training-parameters}

We train all of our networks using the Adam method for stochastic optimization~\cite{kingma2014adam}, with $\beta_1=0.5$ for all trained variables, $\beta_2=0.9$ for our Boltzmann machines, and $\beta_2=0.999$ for our generator and discriminator. As in previous works involving QBMs~\cite{amin2016quantum,khoshaman2018quantum}, we take all $\varGamma_a=2$. During training on synthetic data, we set the learning rate for both of our Boltzmann machines to $10^{-3}$ and use $k=5$ Gibbs steps. Each Boltzmann machine has $8$ visible units and $2$ hidden units, which are sufficient for approximating the studied Bernoulli distribution. 

When training on the MNIST and CIFAR-10 data sets, we consider Boltzmann machines with $32$ visible units and $8$ hidden units. For consistency, we also set the dimension of the noise distribution for DCGAN to be $32$. We use a learning rate of $2\times 10^{-4}$ for both our generator and discriminator, and a learning rate of $10^{-3}$ for our Boltzmann machines with $k=5$ Gibbs steps. Furthermore, we initialize our weights using Xavier initialization~\cite{glorot2010understanding} and initialize our biases to zero. In order to help the discriminator learn in the early stages of trainig, we use soft and noisy labels where a random number between $0$ and $0.1$ is used instead of $0$ labels (fake images) and a random number between $0.9$ and $1$ is used instead of $1$ labels (real images). Each model is trained for $30$ epochs, where an epoch represents one full pass through the training data.

\section{Performance Evaluation Metrics} \label{sec:metrics}

In this section we describe how various GAN architectures are quantitatively compared. For data sets drawn from a known distribution $p_{\textrm{data}}$, we can evaluate the performance of a GAN by computing the KL divergence between the empirical distribution of generated samples and $p_{\textrm{data}}$. Unfortunately, for most image data sets the underlying distribution is unknown and this method is not applicable. Ideally, one would want humans to judge the quality of generated samples in comparison with the original data. However, as this approach is very subjective and almost always impractical, two alternative metrics have been proposed and successfully used to evaluate GANs. 

\subsection{Inception Score}
\label{sec:inception-score}

Salimans et al.~\cite{NIPS2016_6125} proposed the use of a pre-trained neural network---the Inception-v3 network~\cite{szegedy2016rethinking} trained on the ImageNet data set~\cite{deng2009imagenet}---to assess the quality of the generated images. This metric is called the Inception score~\cite{NIPS2016_6125} and is defined as the average KL divergence between the conditional label distribution $p\left(y\given\bm{x}\right)$ and the marginal distribution $p\left(y\right)$ over generated samples; that is:
\begin{equation}\label{inception}
\begin{split}
\mathop{\mathrm{IS}}\left(\mathcal{G}\right) &= \exp\left(\expval{\KLD{p\left(y\given\bm{x}\right)}{p\left(y\right)}}_{\bm{x}}\right)\\
&= \exp\left(S\left(y\right) - \expval{S\left(y\given\bm{x}\right)}_{\bm{x}}\right),
\end{split}
\end{equation}
where $S$ denotes the entropy and the expectation value $\expval{\cdot}_{\bm{x}}$ is taken over image vectors $\bm{x}$ sampled from $\mathcal{G}$. We recognize the term on the right-hand side as the exponential of the mutual information $I(y; \bm{x})$. This metric captures two key features that we are looking for in our generated images: the depiction of clearly identifiable objects (i.e. $p\left(y\given\bm{x}\right)$ should have low entropy for easily classifiable samples) and a high diversity in samples (i.e. $p\left(y\right)$ should have high entropy if all classes are approximately equally represented)~\cite{barratt2018note}.

Note that the use of the Inception network in computing the above score is only appropriate for colored images, which is not the case for the MNIST data set. We follow the approach described in~\cite{kodali2018convergence} and train a simple 4-layer CNN on MNIST, which achieves an accuracy of $99\%$ and therefore can be viewed as a reliable classifier. We then use it to compute the Inception score in an otherwise identical fashion.

The Inception score is a widely adopted evaluation scheme and is known to match well with the human perception of image quality~\cite{NIPS2016_6125}. However, it also has some known drawbacks, such as favoring models that memorize the training data or those that generate clear yet unnatural combinations of objects~\cite{borji2018pros,barratt2018note}.

\subsection{Fr\'{e}chet Inception Distance}
\label{sec:fid}

The Fr\'{e}chet Inception distance, introduced by Heusel et al.~\cite{heusel2017}, avoids some of the problems of the Inception score described above by directly comparing the statistics of synthetic samples to those of real world data. This metric computes the similarity between the features extracted by the pool3 layer of Inception-v3 when the network is supplied with real data and with images generated from $\mathcal{G}$. These features can be thought of as drawn from multivariate Gaussian distributions $\mathcal{N}\left(\bm{\mu}_{\textrm{data}}, \bm{\varSigma}_{\textrm{data}}\right)$ and $\mathcal{N}\left(\bm{\mu}_{\mathcal{G}}, \bm{\varSigma}_{\mathcal{G}}\right)$ respectively. The Fr\'{e}chet distance, also known as the Wasserstein-2 distance, is defined as:
\begin{equation}\label{fid}
\begin{split}
\mathop{\mathrm{FID}}\left(\mathcal{G}\right) &= \norm{\bm{\mu}_{\textrm{data}} - \bm{\mu}_{\mathcal{G}}}^2 \\ &+ \Tr\left(\bm{\varSigma}_{\textrm{data}} + \bm{\varSigma}_{\mathcal{G}} -2\left(\bm{\varSigma}_{\textrm{data}} \bm{\varSigma}_{\mathcal{G}} \right)^{1/2} \right).
\end{split}
\end{equation}
This distance is lower when the features extracted from generated data are distributed similarly to those extracted from real images. 
\end{document}